\pdfoutput=1
\documentclass[11pt]{article}

\usepackage[preprint]{acl}

\usepackage{times}
\usepackage{latexsym}

\usepackage[T1]{fontenc}
\usepackage[utf8]{inputenc}

\usepackage{microtype}

\usepackage{inconsolata}

\usepackage{graphicx}
\usepackage{lipsum}
\usepackage{enumitem}
\usepackage{amsmath}
\usepackage{amssymb}
\usepackage{mathtools}
\usepackage{amsthm}
\newcommand{\ie}{\textit{i}.\textit{e}.}
\newcommand{\eg}{\textit{e}.\textit{g}.}
\usepackage{multirow}
\usepackage{caption}
\usepackage{subcaption}
\usepackage{comment}
\usepackage{mathrsfs}
\usepackage{booktabs}
\usepackage{fontawesome}

\definecolor{aliceblue}{rgb}{0.94, 0.97, 1.0}

\usepackage{booktabs}
\usepackage{colortbl}  %彩色表格需要加载的宏包
\usepackage{xcolor}
\usepackage{pifont}
\usepackage{amsfonts}
\definecolor{aliceblue}{rgb}{0.94, 0.97, 1.0}
\definecolor{deeppink}{RGB}{255,20,147}
\definecolor{mygray}{gray}{0.95}

\usepackage{array} % Important 
\newcolumntype{x}[1]{>{\centering\arraybackslash}p{#1pt}}

\newlength\savewidth

\usepackage{xspace}
\makeatletter
\DeclareRobustCommand\onedot{\futurelet\@let@token\@onedot}
\def\@onedot{\ifx\@let@token.\else.\null\fi\xspace}

\def\eg{\emph{e.g}\onedot} 
\def\ie{\emph{i.e}\onedot} 
\def\cf{\emph{c.f}\onedot} 
\def\etc{\emph{etc}\onedot} \def\vs{\emph{vs}\onedot}

\newcommand{\inlineMarker}{%
  \begingroup\normalfont
  \includegraphics[height=\fontcharht\font`\B]{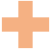}%
  \endgroup
}

\makeatother

\title{\includegraphics[width=0.03\textwidth]{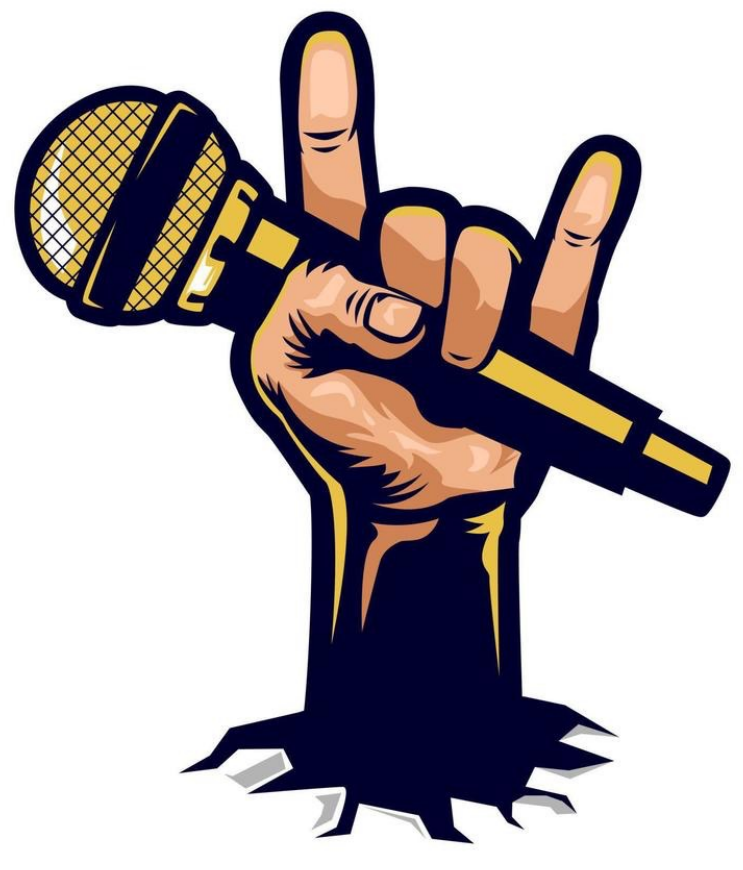} RAP: Efficient Text-Video Retrieval with \\Sparse-and-Correlated Adapter}
\author{Meng Cao\textsuperscript{1,3}$^{*}$, Haoran Tang\textsuperscript{1,2}\thanks{\ \ Equal contributions.}, Jinfa Huang\textsuperscript{1}, Peng Jin\textsuperscript{1,2}, Can Zhang\textsuperscript{1}, Ruyang Liu\textsuperscript{1,2}, \\ \textbf{Long Chen\textsuperscript{4}}, \textbf{Xiaodan Liang\textsuperscript{5,3}}, \textbf{Li Yuan\textsuperscript{1,2}}\thanks{\ \ Corresponding author.}, \textbf{Ge Li\textsuperscript{1}}\\
\textsuperscript{1}School of Electronic and Computer Engineering, Peking University\\ 
\textsuperscript{2}Peng Cheng Laboratory 
\textsuperscript{3}Mohamed bin Zayed University of Artificial Intelligence \\
\textsuperscript{4}The Hong Kong University of Science and Technology 
\textsuperscript{5}Sun Yat-sen University 
}
\begin{document}
\maketitle

\begin{abstract}
Text-Video Retrieval (TVR) aims to align relevant video content with natural language queries. To date, most state-of-the-art TVR methods learn image-to-video transfer learning based on large-scale pre-trained vision-language models (\eg, CLIP). However, fully fine-tuning these pre-trained models for TVR incurs prohibitively expensive computation costs. To this end, we propose to conduct efficient text-video \textbf{R}etrieval with a sparse-and-correlated \textbf{A}da\textbf{P}ter (\textbf{RAP}), \ie, fine-tuning the pre-trained model with a few parameterized layers. To accommodate the text-video scenario, we equip our RAP with two indispensable characteristics: temporal \emph{sparsity} and \emph{correlation}. Specifically, we propose a low-rank modulation module to refine the per-image features from the frozen CLIP backbone, which accentuates salient frames within the video features while alleviating temporal redundancy. Besides, we introduce an asynchronous self-attention mechanism that first selects the top responsive visual patches and augments the correlation modeling between them with learnable temporal and patch offsets. Extensive experiments on four TVR datasets demonstrate that RAP achieves superior or comparable performance compared to the fully fine-tuned counterpart and other parameter-efficient fine-tuning methods.
\end{abstract}
%-----------------------------------------------------------------------
\section{Introduction} \label{sec:intro}
%-----------------------------------------------------------------------
Text-Video Retrieval (TVR) \cite{gabeur2020multi,gorti2022x,he2021improving,lei2021less,luo2022clip4clip,ma2022x,wang2022disentangled} is a pivotal task in the realm of multimodal research, which aims to find the most relevant video content within a repository in response to the text query, and vice versa. With the rapid progress in large-scale image-text pre-training \cite{jia2021scaling,radford2021learning,yu2022coca,yuan2021florence}, current research focuses on how to transfer pre-trained image-text models (\eg, CLIP \cite{radford2021learning}) to the video-text domain. However, fully fine-tuning the video model is computationally expensive and may have the risk of overfitting.

%------------------------------------%
%\begin{comment}
%protect
\begin{figure}[t]
	\centering
	\includegraphics[width=0.5\textwidth]{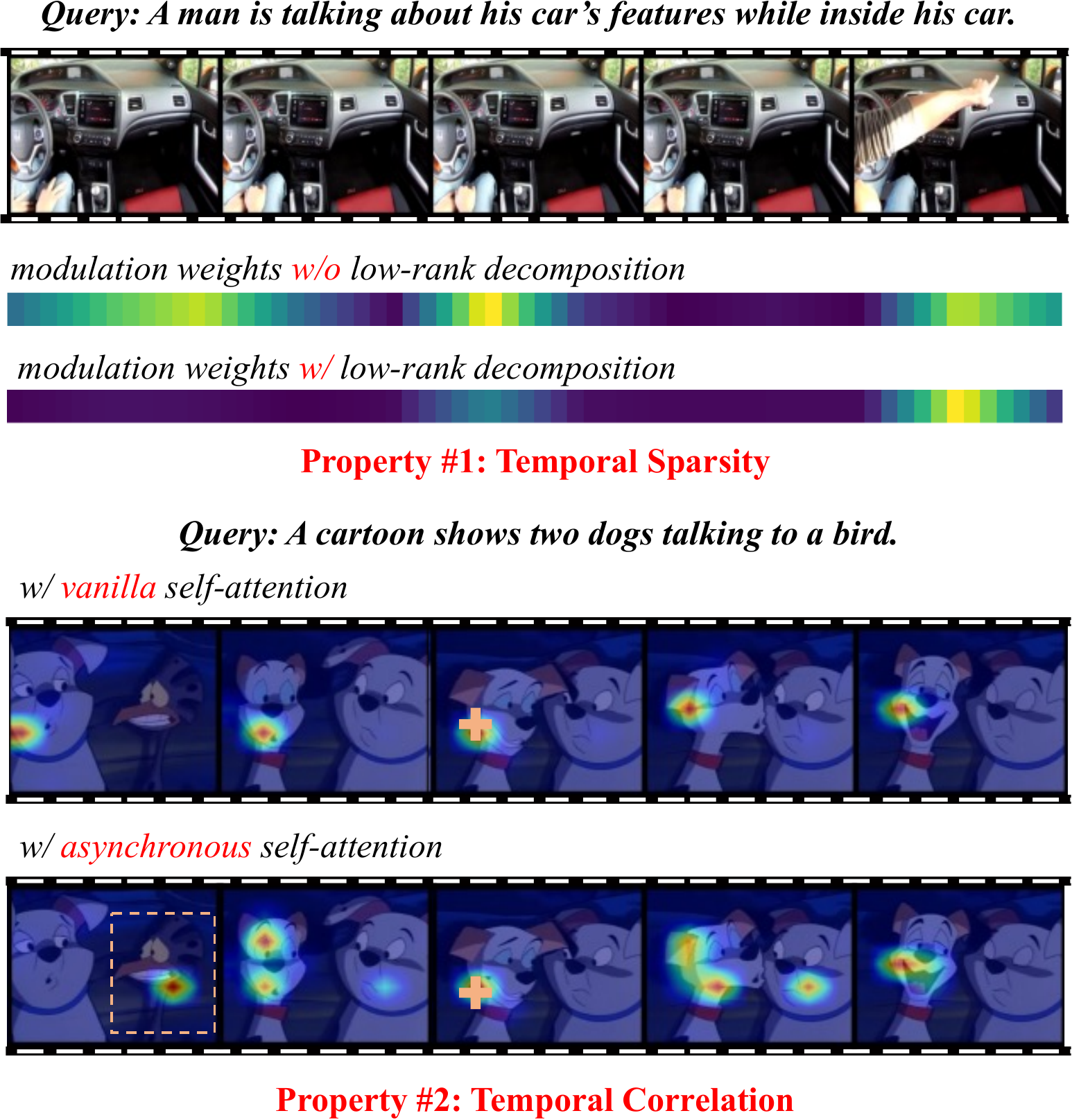}
	\caption{\textbf{Top: Illustrations of temporal sparsity.} We visualize the modulation weight \emph{w/} or \emph{w/o} low-rank decomposition. \textbf{Down: Illustrations of temporal correlation.} The query patch is marked by the yellow cross and the similarity map within other frames are plotted.}
	\label{fig:teaserMotivate}
	%\vspace{-2mm}
\end{figure}

\begin{figure}[t]
	%\left
\includegraphics[width=0.46\textwidth]{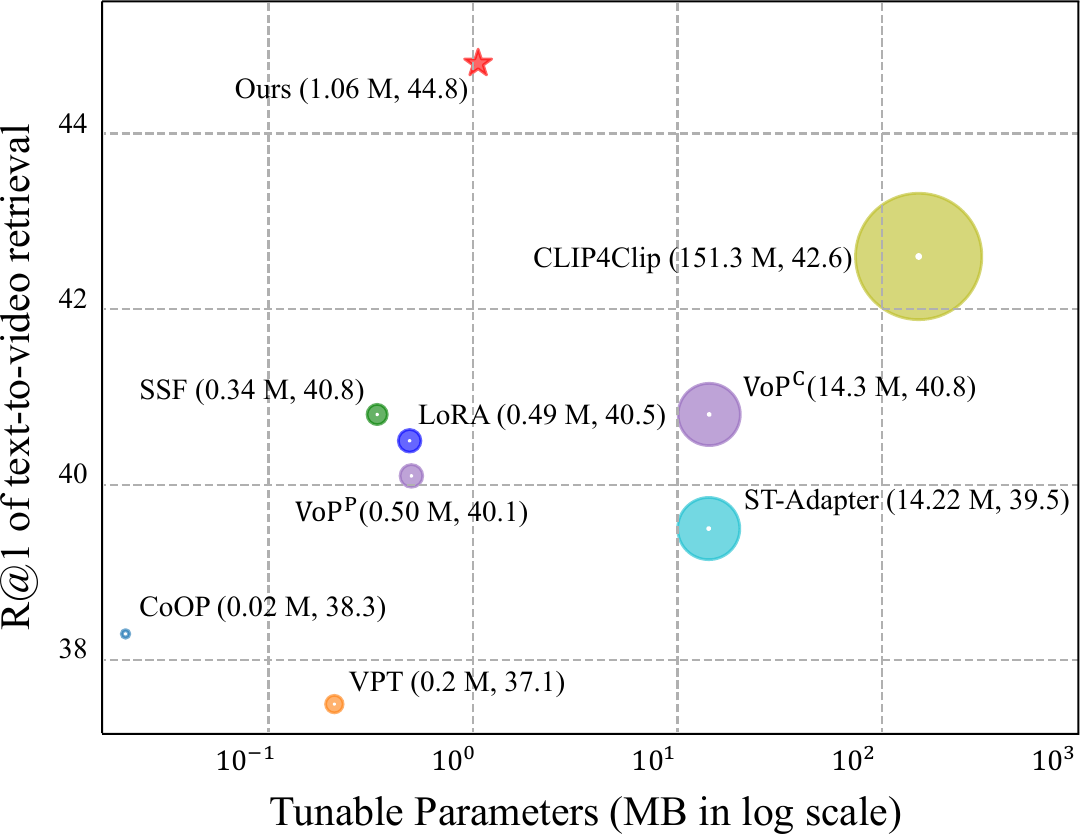}
\caption{\textbf{Text-to-video retrieval performance on MSR-VTT dataset.} Marker sizes are proportional to the number of tunable parameters.}
\label{fig:teaserCompare}
%\vspace{-2mm}
\end{figure}
%------------------------------------%

To alleviate this dilemma, Parameter-Efficient Fine-Tuning (PEFT) stemmed from natural language processing \cite{houlsby2019parameter,lester2021power,zaken2022bitfit,hu2021lora} has also aroused extensive research interest in the field of computer vision \cite{chen2022vision,chen2022adaptformer} and cross-modal learning \cite{sung2022vl}. Recently, some exploratory work \cite{zhang2023multimodal,jiang2022cross,diao2024unipt} has also attempted to introduce PEFT into TVR. These methods, however, simply introduce existing PEFT algorithms \cite{houlsby2019parameter,you2022learning,karimi2021compacter} without considering the inherent characteristics of video data.

To this end, we argue that an ideal PEFT method for VTR should be equipped with two characteristics: 1) \textbf{Temporal Sparsity}: As shown in Figure~\ref{fig:teaserMotivate}, the video data inherently contains lots of redundancies or repetitions in the temporal perspective. The visualized frame-by-frame embedded CLIP features are \emph{over-smooth}, resulting in the loss of important details or nuances within the video data. In contrast, the video feature adapted from pre-trained CLIP should capture the most informative frames, allowing for a more sparse representation. 2) \textbf{Temporal Correlation}: The desired video adapter is supposed to incorporate the dependencies and relationships between consecutive frames, especially when dealing with actions or events that happen in several frames, as the features can encapsulate the evolving context over time. For example in Figure~\ref{fig:teaserMotivate}, the query sentence contains two entities including \texttt{dog} and \texttt{bird}. Given the query patch (\inlineMarker \ in frame \#3), we visualize the similarity distribution within the other patches. As in this example, vanilla self-attention can only attend to the \texttt{dog} instance while the other \texttt{bird} instance is overlooked.

In the realm of video processing and analysis, the temporal dimension often contains redundancies due to the inherent correlation between adjacent frames. This redundancy can lead to inefficiencies in computational resources and storage when dealing with large-scale video data. Therefore, there is a need to extract meaningful and informative features while reducing temporal redundancy.

To alleviate these aforementioned issues, we propose an efficient text-video \textbf{R}etrieval framework with sparse-and-correlated \textbf{A}da\textbf{P}ter (dubbed as \textbf{RAP}). Our proposed RAP not only streamlines the trainable parameters, enhancing efficiency in computational resources, but also tailors the architecture to adeptly capture and model the nuanced temporal characteristics of video data.

To achieve temporal sparsity, we propose a Low-Rank Modulation (\textbf{LoRM}) module to refine the pre-trained CLIP feature \cite{radford2021learning} on the principle of redundancy reduction and essential information extraction. This design stems from a simple hypothesis that the change in temporal weights resides on a low intrinsic rank \cite{zhang2012slow}. Therefore, we introduce layer-wise low-rank scale parameters and shift parameters, which could be considered as variance and mean to modulate the CLIP feature. Specifically, both scale and shift parameters are instantiated by the multiplication of two low-rank trainable matrices. These parameters are input-independent and therefore more flexible. LoRM allows us to calibrate the video features to highlight salient frames and mitigate temporal redundancy.

For temporal correlation modeling, we replace vanilla self-attention with the proposed Asynchronous Self-Attention (\textbf{ASA}), which introduces temporal dynamics among video frames to capture temporal relationships. Since the attention computing in pre-trained CLIP is constrained within each frame feature, it is challenging to apply to the video domain due to the temporally dynamic nature of video frames. Previous methods employ either temporal Transformer \cite{jiang2022cross,yang2022aim,zhang2023multimodal,cao2021unifacegan} or 3D convolution networks \cite{yao2023side4video,liu2023revisiting} to encode temporal dependencies. Instead of introducing additional modules, we propose an asynchronous self-attention that only warps partial patch tokens in a parameterized way. Firstly, for each frame, we filter semantically significant patches via a parameter-free text-conditioned selection mechanism. Specifically, we compute the similarities between patch features and the corresponding sentence and select the patches with the highest responses. Secondly, each selected patch within the current frame is dynamically warped to attend to the temporally related patches in other frames. The proposed asynchronous self-attention empowers the flexibility in capturing correlations between video frames at the fine-grained patch level.

Overall, the main contributions of this work are:

\begin{itemize}[topsep=0pt, partopsep=0pt, leftmargin=13pt, parsep=0pt, itemsep=3pt]
    \item We propose RAP to adapt the pre-trained CLIP to efficient TVR, which not only reduces the tunable parameters but also generates temporally sparse and correlated video features.

    \item To alleviate the temporal redundancy, a low-rank modulation module is introduced to calibrate the frame-wise representation linearly.

    \item We propose an asynchronous self-attention that captures long-range dependencies with negligible computational overheads.
	
    \item Extensive experiments show that our RAP is on par with or even superior to previous PEFT methods and the fully fine-tuned counterpart.    
\end{itemize}

%-----------------------------------------------------------------------
\section{Related Work}\label{relatedwork}
%-----------------------------------------------------------------------

\noindent \textbf{Text-Video Retrieval.}  TVR \cite{yu2018joint,croitoru2021teachtext,yang2021taco,wang2021t2vlad,chen2020fine,wang2023video,jin2022expectation,jin2023video,jin2023diffusionret,liu2022ts2} is a fundamental research topic in the video-language domain aiming to retrieval the relevant video/text based on the given text/video query, which showcases significant practical potentials in video grounding \cite{li2024exploiting,li2023g2l,li2023exploiting,cao2023iterative,cao2022deep,cao2021pursuit,zhang2021cola,zhang2021synergic}, referring expression comprehension \cite{ji2023video,cao2022correspondence,jiang2022video,cao2021all}, video caption \cite{yang2023concept,mao2023improving,ji2022visual,ye2023qilin,liu2023qilin} , \etc. The pioneer works \cite{yu2018joint,gabeur2020multi} rely on pre-extracted features from frozen video and text encoders. To facilitate the end-to-end training, ClipBERT \cite{lei2021less} proposes a sparse sampling strategy for efficient text-video training. With the great success of large-scale image-text pretraining model CLIP \cite{radford2021learning}, the majority of the state-of-the-art TVR methods \cite{luo2022clip4clip,ma2022x,wang2023unified,hannan2023rgnet,jin2022expectation} focus on transferring the powerful CLIP encoder to the video-text domain by designing various cross-modal alignment strategies. As the first attempt, CLIP4Clip \cite{luo2022clip4clip} employs mean-pooling or Transformer to aggregate video features and conduct coarse-grained (video-sentence level) contrastive alignment. Instead of using the text-agnostic aggregation manner, X-CLIP \cite{ma2022x} proposes to aggregate video representations conditioned on the text’s attention weight and conduct the multi-grained contrastive learning at the frame-word, video-sentence, video-word and sentence-frame levels. For more comprehensive alignment, UCOFIA \cite{wang2023unified} unifies the coarse-grained and fine-grained alignment to capture both the high-level and low-level correspondence between text and video.

Most of the current TVR methods follow the fully fine-tuning paradigm. This scheme, however, is computation-intensive and may have the risk of overfitting. Besides, additional temporal modeling models are required to bridge the image and video gap. In this paper, we propose RAP which conducts parameter-efficient fine-tuning for TVR which provides a more computationally efficient and potentially more robust approach. Besides, the tunable parameters in our RAP also bear the responsibility for temporal modeling, thus eliminating the need for external temporal modules.

\noindent \textbf{Parameter-Efficient Transfer Learning.} PEFT \cite{houlsby2019parameter,hu2021lora,lester2021power,he2021towards,zaken2022bitfit,sung2021training} is firstly introduced in the NLP domain to reduce trainable parameters while maintaining the comparable performance with the fully fine-tuning setting. Inheriting the merit from NLP, PEFT in computer vision \cite{jia2022visual,bahng2022visual,jie2022convolutional,sung2022vl} also gained extensive research attention. VPT \cite{jia2022visual} follows the prompt tuning strategy by introducing the task-specific learnable prompts on vision Transformer. To be compatible with vision tasks, Convpass \cite{jie2022convolutional} introduces the inductive bias of convolutional layers by reconstructing the spatial structure of the token sequence via convolution operations. VL-Adapter \cite{sung2022vl} pioneeringly benchmarks different types of PEFT techniques including Adapter \cite{houlsby2019parameter}, Hyperformer \cite{mahabadi2021parameter}, and Compacter \cite{karimi2021compacter} in the multitask setting. 

There also exist several works \cite{yang2022aim,pan2022st,lin2022frozen,li2023zeroi2v,yao2023side4video,jiang2022cross,zhang2023multimodal,lu2023uniadapter,cao2022locvtp,zhang2022unsupervised} focusing on the image-to-video transfer learning. Based on the pre-trained CLIP model, these methods either introduce temporal convolution \cite{pan2022st} or Transformer \cite{lu2023uniadapter} in sequential \cite{zhang2023multimodal,jiang2022cross} or parallel \cite{yao2023side4video} ways. However, they overlook the inherent temporal structure of video data while our RAP pinpoints two key issues in video feature modeling and generate more representative video features.

\begin{figure*}[t]
\centering
\includegraphics[width=\textwidth]{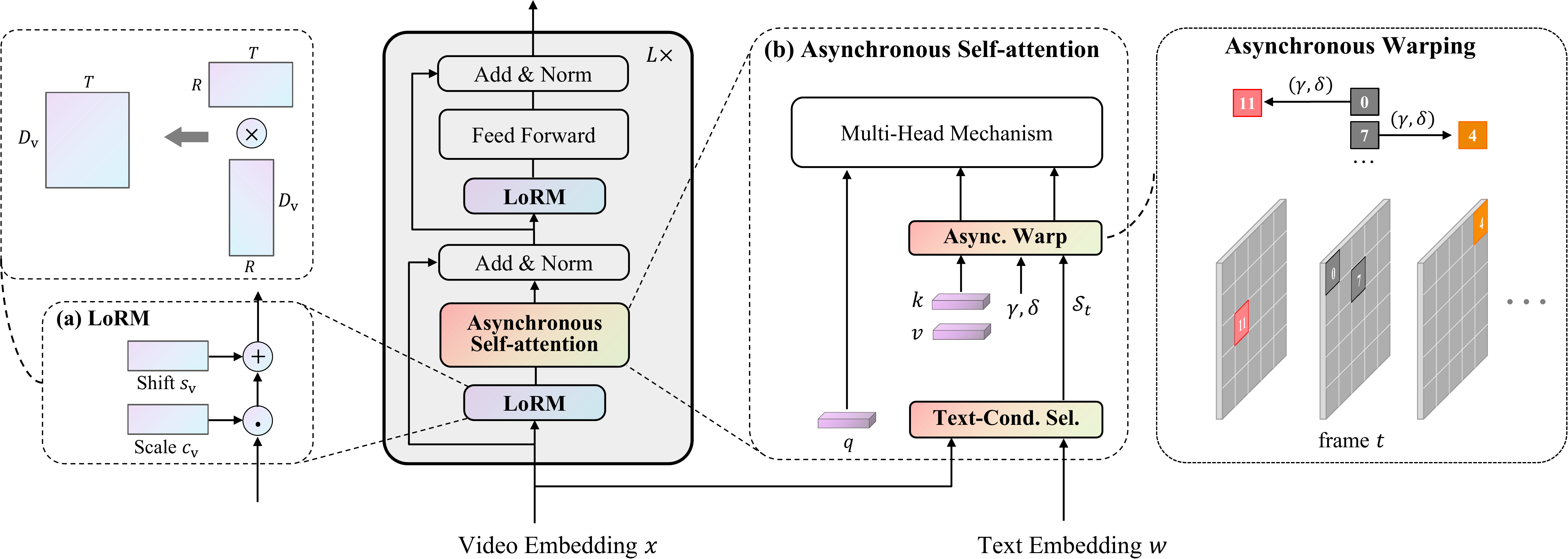}
\caption{\textbf{An overview of RAP.} (a) LoRM sets up learnable shift parameters $\boldsymbol{c}_\text{v}$ and scale parameters $\boldsymbol{s}_\text{v}$  to calibrate the vanilla CLIP features. For the temporally sparse requirement, $\boldsymbol{c}_\text{v}$ and $\boldsymbol{s}_\text{v}$ are generated by low-rank decomposition on the temporal dimension. (b) Asynchronous self-attention first filters out patch set $\mathcal{S}_t$ via text-conditioned selection. Then, the filtered patches are warped based on the learnable patch offset $\gamma$ and temporal offset $\delta$.}
\label{fig:pipeline}
%\vspace{-2mm}
\end{figure*}
%-----------------------------------------------------------------------
\section{Method} \label{method}
%-----------------------------------------------------------------------

Text-video retrieval aims to search for and retrieve relevant videos/texts based on textual/video queries by evaluating the similarity between the video-sentence pairs. Our proposed RAP is devoted to bridging the gap between the frozen CLIP feature and the dynamic video scenario by introducing negligible parameter overheads. 

The schematic illustration of our RAP is illustrated in Figure \ref{fig:pipeline}. In Sec.~\ref{sec:3.1}, we first present the preliminaries of RAP including the video and text feature embedding. Then we describe the proposed low-rank modulation and the asynchronous self-attention in Sec.~\ref{sec:3.2} and Sec.~\ref{sec:3.3}, respectively. 

%-----------------------------------------------------------------------
\subsection{Feature Embedding} \label{sec:3.1}
%-----------------------------------------------------------------------

%-----------------------------------------------------------------------
\noindent \textbf{Video Embedding.} We utilize the visual backbone (ViT \cite{dosovitskiy2020image}) of CLIP as the video encoder. Given the video data, we follow ViT \cite{dosovitskiy2020image} to process each frame independently. Specifically, each frame with shape $H\times W$ is split into non-overlapping patches with shape $P\times P$ and then linearly projected into the embedding space. Such linear projection generates $N = H W / P^2$ patch features for each frame. Besides, a learnable \texttt{[CLS]} token is prepended to each frame patch feature sequence to represent the global frame representations. The positional embedding is also added to incorporate positional information explicitly. Through the above process, we obtain the $t^\text{th}$ frame feature $\boldsymbol{x}^{0}_{t} \in \mathbb{R}^{(N+1) \times D_\text{v}}$, $t \in [1, T]$, where $D_\text{v}$ is visual feature dimension and $T$ is the total frame number.

The residual structure with serially connected multi-head self-attention (MHSA) and multilayer perceptron (MLP) is applied to capture sequential dependencies and contextual relationships within each frame patch sequence. Repeating the above steps for each frame, we obtain the video embedding at $l^\text{th}$ layer $\boldsymbol{x}^{l} \in \mathbb{R}^{T \times (N + 1) \times D_\text{v}}$, $l \in [1, L]$, where $L$ denotes the layer number. Specifically, we decompose $\boldsymbol{x}^{l} = [\boldsymbol{f}^{l}, \boldsymbol{p}^{l}]$, where $\boldsymbol{f}^{l} \in \mathbb{R}^{T \times D_\text{v}}$ represents the frame-wise features (\ie, \texttt{[CLS]} token feature) while $\boldsymbol{p}^{l} \in \mathbb{R}^{T \times N \times D_\text{v}}$ is patch-wise representation at the $l^\text{th}$ layer.

\noindent \textbf{Text Embedding.} For text embedding, we directly use the text encoder of CLIP to generate the textual representation. The text encoder is a Transformer \cite{vaswani2017attention} with the architecture modifications as described in \cite{radford2019language}. An \texttt{[EOS]} token is also appended to encode the global sentence feature. Concretely, we denote the sentence features at the $l^\text{th}$ layer as $\boldsymbol{w}^{l} \in \mathbb{R}^{1 \times D_\text{t}}$, where $D_\text{t}$ is the text feature dimension.
%-----------------------------------------------------------------------

%-----------------------------------------------------------------------
\subsection{Low-rank Modulation} \label{sec:3.2}
%-----------------------------------------------------------------------

%-----------------------------------------------------------------------
In this section, we elaborate on the feature modulation for both video and text features. Since all the layers share the same modulation process, we omit the superscript of layer index $l$ for brevity.

\noindent \textbf{Low-rank Modulation for Video.} The frame-by-frame encoded video features $\boldsymbol{x}$ cannot reflect the characteristics of the video data. The redundancy in the temporal dimension is a major feature that distinguishes videos from static images. To this end, we introduce low-rank scale parameters and shift parameters, which serve as the variance and mean values to modulate the pre-trained CLIP feature. These parameters are input-independent, rendering them comparatively lightweight in nature and hopefully more scalable. Specifically, the video scale parameter $\boldsymbol{c}_\text{v} \in \mathbb{R}^{T \times D_\text{v}}$ and video shift parameter $\boldsymbol{s}_\text{v} \in \mathbb{R}^{T \times D_\text{v}}$ are decomposed as follows:
\begin{equation}
\boldsymbol{c}_\text{v} = \boldsymbol{c}^{\text{a}} \cdot \boldsymbol{c}^{\text{b}}, \enspace \boldsymbol{s}_\text{v} = \boldsymbol{s}^{\text{a}} \cdot \boldsymbol{s}^{\text{b}}, \label{eq:1}
\end{equation}
\noindent where $\boldsymbol{c}^{\text{a}}, \boldsymbol{s}^{\text{a}} \in \mathbb{R}^{T \times R}$, $\boldsymbol{c}^{\text{b}}, \boldsymbol{s}^{\text{b}} \in \mathbb{R}^{R \times D_\text{v}}$ are learnable parameters and we set rank $R \ll \min(T, D_\text{v})$ to implement the low-rank requirement. The low-rank modulation is applied as follows.
\begin{equation}
\boldsymbol{u} = \boldsymbol{c}_\text{v} \odot \boldsymbol{x} + \boldsymbol{s}_\text{v}, \label{eq:2}
\end{equation}
\noindent where $\odot$ denotes the element-wise multiplication with broadcast. During training, the vanilla feature $\boldsymbol{x}$ is extracted through frozen CLIP backbone and the learnable $\boldsymbol{c}_\text{v}$ and $\boldsymbol{s}_\text{v}$ help modify $\boldsymbol{x}$ to be of temporally low-rank. $\boldsymbol{u} \in \mathbb{R}^{T \times (N + 1) \times D_\text{v}}$ is the modulated video feature. 

\noindent \textbf{Modulation for Text.} We also modulate the textual embedding $\boldsymbol{w}$ with parameters $\boldsymbol{c}_\text{t}$ and $\boldsymbol{s}_\text{t}$ as follows.
\begin{equation}
\boldsymbol{z} = \boldsymbol{c}_\text{t} \odot \boldsymbol{w} + \boldsymbol{s}_\text{t},
\end{equation}
\noindent where $\boldsymbol{c}_\text{t}, \boldsymbol{s}_\text{t} \in \mathbb{R}^{1 \times D_\text{v}}$ are learnable parameters. We do \textbf{not} conduct modulation at the word level or use parameter low-rank decomposition since the textual data do not exhibit the sparsity characteristic. 
%-----------------------------------------------------------------------

%-----------------------------------------------------------------------
\subsection{Asynchronous Self-Attention} \label{sec:3.3}
%-----------------------------------------------------------------------

%-----------------------------------------------------------------------
Let's review the vanilla self-attention in the video encoder. For clarity, we take the $t^{\text{th}}$ frame of the input video for illustration. The corresponding modulated feature is denoted as $\boldsymbol{u}_t \in \mathbb{R}^{N \times D_\text{v}}$, $t \in [1, T]$ (\cf. Equation~\eqref{eq:2}). Note that here we define $\boldsymbol{u}_t$ as the patch-wise feature which does not contain the global \texttt{[CLS]} token features. We also omit the superscript of layer index $l$.

The vanilla self-attention first performs three different linear projections on the input feature $\boldsymbol{{u}}_t$ to obtain the triplet of query, key, and value, \ie,
\begin{equation}
\boldsymbol{q}_{t} = \boldsymbol{u}_t \cdot \mathbf{W}_{q}, \; \boldsymbol{k}_{t} = \boldsymbol{u}_t \cdot \mathbf{W}_{k}, \; \boldsymbol{v}_{t} = \boldsymbol{u}_t \cdot\mathbf{W}_{v}, 
\label{eq:4}
\end{equation}
\noindent where $\mathbf{W}_{q}, \mathbf{W}_{k}, \mathbf{W}_{v} \in \mathbb{R}^{D_\text{v} \times D_\text{v}}$ are frozen transformation weights. Then the scaled dot-product attention is computed to achieve the contextual information.

The vanilla self-attention only attends to the intra-frame correlation modeling, which leads to the modality gap between video and image. Instead of introducing an additional serial or parallel temporal modeling module (temporal Transformer \cite{liu2023revisiting,yang2022aim} or 3D Convolution \cite{pan2022st}), we propose a novel asynchronous self-attention which introduces patch-wise temporal offset to model inter-frame relationship. Besides, to stabilize the training process, we propose a text-conditioned selection mechanism. 

% Version 1
\noindent \textbf{Text-conditioned Selection.} Here we take the video-to-text retrieval as an example to illustrate this. For given frame-wise video feature $\boldsymbol{f} \in \mathbb{R}^{T \times D_\text{v}}$, we conduct mean pooling on the frame dimension to obtain the video-level features $\boldsymbol{\overline{f}} \in \mathbb{R}^{1 \times D_\text{v}}$. Then we select the most similar sentence $\boldsymbol{w}^{*} \in \mathcal{W}$ as follows.
\begin{equation}
\boldsymbol{w}^{*} =\underset{\boldsymbol{w} \in \mathcal{W}}{\arg\max}\left(\operatorname{Proj}(\boldsymbol{\overline{f}}) \cdot \boldsymbol{w}^\intercal \right),
\end{equation}
\noindent where $\boldsymbol{w} \in \mathbb{R}^{1 \times D_\text{t}}$ is the candidate sentence features. $\operatorname{Proj}(\cdot)$ is a linear projection layer to transform the visual dimension $D_\text{v}$ to the textual dimension $D_\text{t}$.

Then, we compute the sentence-patch similarity and select the top $K$ responded patches.
\begin{equation}
\mathcal{S}_t =\underset{t \in\left[1, T\right]}{\arg \operatorname{topk}}\left(\operatorname{Proj}(\boldsymbol{u}_{t}) \cdot \boldsymbol{w}^{*\intercal}\right),
\end{equation}
\noindent where $\mathcal{S}_t$ is the filtered patch index set. 

\noindent \textbf{Asynchronous Self-Attention.} Then we only apply the proposed asynchronous self-attention on patches indexed by the set of $\mathcal{S}_t$. Specifically, the query features are adapted as follows.
\begin{equation}
\hat{\boldsymbol{k}}_t^n, \hat{\boldsymbol{v}}_t^n =
\begin{cases}
\boldsymbol{k}_{t + \boldsymbol{\delta}_{t}}^{n + \boldsymbol{\gamma}_{n}}, \boldsymbol{v}_{t + \boldsymbol{\delta}_{t}}^{n + \boldsymbol{\gamma}_{n}}, & n \in \mathcal{S}_t \\
\boldsymbol{k}_t^n, \boldsymbol{v}_t^n, & n \notin \mathcal{S}_t
\end{cases}
\label{eq:7}
\end{equation}
\noindent where $\boldsymbol{\gamma} \in \mathbb{R}^{N \times 1}$, $\boldsymbol{\delta} \in \mathbb{R}^{T \times 1}$ are layer-shared learnable parameters representing the offset distance in the patch and temporal dimension, respectively. $\boldsymbol{k}_{t + \boldsymbol{\delta}_{t}}^{n + \boldsymbol{\gamma}_{n}}$ and $\boldsymbol{v}_{t + \boldsymbol{\delta}_{t}}^{n + \boldsymbol{\gamma}_{n}}$ denote the key and value features of the $(n+\gamma_n)^\text{th}$ patch in the $(t+\delta_t)^\text{th}$ frame, respectively. $\boldsymbol{\hat{k}}_{t}, \boldsymbol{\hat{v}}_{t} \in \mathbb{R}^{N \times D_\text{v}}$ represents the adapted features. Finally, asynchronous self-attention is computed as follows.
\begin{equation}
\operatorname{Atten}(\boldsymbol{{q}}_{t}, \boldsymbol{\hat{k}}_{t}, \boldsymbol{\hat{v}}_{t})=\operatorname{softmax}(\frac{\boldsymbol{q}_{t} \boldsymbol{\hat{k}}_{t}^\intercal}{\sqrt{D_\text{v}}}) \boldsymbol{\hat{v}}_{t},
\end{equation}
\noindent where $\boldsymbol{q}_{t}$ is illustrated in Equation \eqref{eq:4} while $\boldsymbol{\hat{k}}_{t}$ and $\boldsymbol{\hat{v}}_{t}$ are defined in Equation \eqref{eq:7}.
%-----------------------------------------------------------------------

%-------------------------------------------------------------------------------------%
\section{Experiments}\label{expes}
%-------------------------------------------------------------------------------------%

%We first introduce the experimental settings in Sec.~\ref{sec:4_1}. The comparisons with the state-of-the-art methods are discussed in Sec.~\ref{sec:4_2}. Then, we present the model architecture agnostic analysis in Sec.~\ref{sec:4_3}. The extensive exploratory studies on cross-modal correspondence and inter-frame correspondence learning are provided in Sec.~\ref{sec:4_4} and Sec.~\ref{sec:4_5}, respectively. Note that all ablation studies are conducted on the VID-Sentence dataset. Further, we provide the visualization results in Sec.~\ref{sec:4_6}.

%-------------------------------------------------------------------------------------%
%-------------------------------------------------------------------------------------%
\begin{table*}[t]
\centering
\renewcommand\arraystretch{1.1}
\vspace{-1em}
\caption{\textbf{Comparisons with state-of-the-art methods on MSR-VTT dataset.} \faLock~ denotes using the frozen visual encoder. RAP$^{*}$ denotes the RAP model with DSL post-processing \cite{cheng2021improving}. \faUnlock~refers to the text-encoder being trainable. The best performance is in \textbf{bold} and the second best is \underline{underlined}.}
\vspace{-5pt}
\resizebox{\linewidth}{!}
{
%\begin{tabular}{llcllllllll}
\begin{tabular}{llccccccccc}

\toprule%[1.5pt]
\multicolumn{2}{c}{} & Trainable & 
\multicolumn{4}{c}{Text → Video} & \multicolumn{4}{c}{Video → Text} \\
Type & Methods &  Params (MB) ↓ & R@1↑ & R@5↑ & R@10↑ & MnR↓ & R@1↑ & R@5↑ & R@10↑ & MnR↓ \\ 
\midrule
\rowcolor{mygray}\multicolumn{11}{l}{\emph{with CLIP-ViT-B/32}} \\
\multicolumn{1}{l}{\multirow{2}{*}{Fine-tune}} &
  \multicolumn{1}{l}{CLIP4Clip \cite{luo2022clip4clip}} & 151.28 & \multicolumn{1}{|l} {42.6} &  70.8 & 79.9 & \multicolumn{1}{l|}{16.1} & 43.9 & 70.0 & 81.4 & 11.7 \\ 
& \multicolumn{1}{l}{CLIP4Clip (\faLock~CLIP)} & 0 & \multicolumn{1}{|l}{31.1} & 53.7 & 63.4 & \multicolumn{1}{l|}{41.6} & 26.5 & 50.1 & 61.7 & 39.9 \\ 
\hline
\multicolumn{1}{l}{\multirow{5}{*}{Prompt}} & \multicolumn{1}{l}{VPT \cite{jia2022visual}} & 0.21 &  
\multicolumn{1}{|l}{37.5} & 63.0 & 73.9 & \multicolumn{1}{l|}{21.6} & 36.5 & 62.8 & 74.3 & 20.0 \\     
& \multicolumn{1}{l}{VPT~\faUnlock ~\cite{jia2022visual}} & 63.43 &  
\multicolumn{1}{|l}{40.5} & 67.3 & 78.6 & \multicolumn{1}{l|}{17.9} & 40.9 & 70.0 & 79.2 & 12.5 \\
\multicolumn{1}{c}{} & \multicolumn{1}{l}{CoOp \cite{zhou2022learning}} & \textbf{0.02} & \multicolumn{1}{|l}{38.3} & 62.3 & 73.4 & \multicolumn{1}{l|}{18.9} & 41.0 & 66.6 & 77.4 & 13.4 \\
\multicolumn{1}{c}{} & \multicolumn{1}{l}{{VoP$^{P}$} \cite{huang2023vop}} & {0.50} & \multicolumn{1}{|l}{40.1} & 65.7 & 77.7 & \multicolumn{1}{l|}{16.9} & 42.5 & 70.0 & 79.9 & 12.4 \\
\multicolumn{1}{c}{} & \multicolumn{1}{l}{{VoP$^{C}$} \cite{huang2023vop}} & 
{14.30} & \multicolumn{1}{|l} {40.8} & 68.1 & 79.0 & \multicolumn{1}{l|}{15.8} & 42.3 & 70.1 & 81.1 & 11.4 \\
\hline 
\multicolumn{1}{l}{\multirow{4}{*}{Adapter}} & \multicolumn{1}{l}{ST-Adapter \cite{pan2022st}} & 14.22 & \multicolumn{1}{|l}{39.5} & 65.1 & 74.2 & \multicolumn{1}{l|}{20.0} & 37.1 & 64.5 & 75.9 & 19.7 \\
& \multicolumn{1}{l}{ST-Adapter~\faUnlock ~\cite{pan2022st}} & 77.45 & \multicolumn{1}{|l}{42.5} & 70.0 & 80.1 & \multicolumn{1}{l|}{17.0} & 42.1 & 70.0 & 81.2 & 11.4 \\
 & \multicolumn{1}{l}{LoRA \cite{hu2021lora}} & 0.49 & \multicolumn{1}{|l}{40.5} & 67.1 & 78.9 & \multicolumn{1}{l|}{16.4} & 42.1 & 70.0 & 79.8 & 13.5 \\
 & \multicolumn{1}{l}{SSF \cite{lian2022scaling}} & 0.34 & \multicolumn{1}{|l}{40.8} & 68.2 & 78.6 & \multicolumn{1}{l|}{17.0} & 42.0 & 68.6 & 80.2 & 13.2 \\ 
 % & \multicolumn{1}{l}{Side4Video \cite{xxx}} & - & 44.2 & 71.1 & 81.0 & 15.1 & 44.6 & 72.3 & 82.3 &  9.4 \\
\rowcolor{aliceblue}  & \multicolumn{1}{l}{{RAP (Ours)}} & 1.06 & \multicolumn{1}{|l}{\textbf{44.8}} & \textbf{71.4} & \textbf{81.5} & \multicolumn{1}{l|}{\textbf{14.4}} & \textbf{44.0} & \textbf{71.9} & \textbf{82.4} & \textbf{10.1} \\
\hline
\rowcolor{mygray}\multicolumn{11}{l}{\emph{with CLIP-ViT-B/16}} \\
\multicolumn{1}{c}{\multirow{6}{*}{}} & \multicolumn{1}{l} {CLIP4Clip \cite{luo2022clip4clip}}  & 149.62 & \multicolumn{1}{|l}{45.4} & 72.1 & 81.1 & \multicolumn{1}{l|}{14.5} & 44.9 & 72.2 & 81.8 & 10.4  \\
& \multicolumn{1}{l} {{VoP$^{P}$} \cite{huang2023vop} }  & 0.50 & \multicolumn{1}{|l}{43.9} & 70.0 & 80.9 & \multicolumn{1}{l|}{12.9} & - & - & - & -  \\
& \multicolumn{1}{l} {{VoP$^{C}$}
\cite{huang2023vop} } & 14.30 & \multicolumn{1}{|l}{44.6} & 71.8 & 80.2 & \multicolumn{1}{l|}{14.6} & - & - & - & -  \\
& \multicolumn{1}{l} {{MV-Adapter} \cite{zhang2023multimodal} }  & 3.87 & \multicolumn{1}{|l}{46.0} & 72.0 & \underline{82.1} & \multicolumn{1}{c|}{-} & \multicolumn{1}{c}{\underline{45.6}} & 74.0 & 83.8 & -  \\
\rowcolor{aliceblue}  & \multicolumn{1}{l}{RAP (Ours) } &{1.06} & \multicolumn{1}{|l}{\underline{46.5}} & \underline{73.9} & 82.0 & \multicolumn{1}{l|}{\underline{12.1}} & 45.3 & \underline{76.4} & \underline{84.8} & \underline{9.1} \\ 
%\rowcolor{gray!30}  &
%\multicolumn{1}{l} {+ DSL\cite{cheng2021improving}}  &
\rowcolor{aliceblue} & {RAP$^{*}$ (Ours)} & 1.06 & \multicolumn{1}{|l}{\textbf{52.1}} & \textbf{77.3} & \textbf{86.7} & \multicolumn{1}{l|}{\textbf{10.0}} &  \textbf{51.6}& \textbf{78.7} & \textbf{86.9} & \textbf{8.0} \\   
\bottomrule %[1.5pt]
\end{tabular}
}
\label{tab:msrvtt}
\end{table*}

\subsection{Experimental Settings}\label{sec:4_1}
%-------------------------------------------------------------------------------------%

%-------------------------------------------------------------------------------------%
\noindent \textbf{Datasets.} We validate the performance of our proposed RAP on four benchmarked datasets. 1) \textbf{MSR-VTT} \cite{xu2016msr} contains 10,000 YouTube videos and each video is associated with 20 textual descriptions. We follow the 1k-A split \cite{yu2018joint} where 9,000 videos are used for training and 1,000 videos for testing. 2) \textbf{MSVD} \cite{chen2011collecting} is composed of 1,970 videos. Following the official split, we used 1,200 videos for training and 670 videos for testing, respectively. 3) \textbf{ActivityNet Captions} \cite{krishna2017dense} covers 20,000 untrimmed videos of complex human activities with an average duration of two minutes. We report results on the ``val1'' split (10,009 training videos and 4,917 testing videos) as in \cite{gabeur2020multi}. 4) \textbf{DiDemo} \cite{anne2017localizing} consists of 10,464 unedited, personal videos in diverse visual settings annotated with 40,543 text descriptions. We follow the training and evaluation protocol in \cite{luo2022clip4clip}.
%-------------------------------------------------------------------------------------%

%-------------------------------------------------------------------------------------%
\noindent \textbf{Evaluation Metrics.} Following the previous work \cite{luo2022clip4clip}, we evaluate the performance with standard retrieval metrics: recall at rank $K$ (R@$K$, higher is better), median rank (MdR, lower is better) and mean rank (MnR, lower is better). R@$K$ is defined as the percentage of samples for which the correct result is found in the top-$K$ retrieved results. We set $K$ to $\{1, 5, 10\}$ in our experiments. MdR calculates the median of the ground-truth results in the ranking while MnR calculates the mean rank of all the correct results.
%-------------------------------------------------------------------------------------%

%-------------------------------------------------------------------------------------%
\noindent \textbf{Implementation Details.} We set the input frame length to 12, 64, 12, 64 and the caption token length to 32, 64, 32, 64 for MSR-VTT, DiDeMo, MSVD, and ActivityNet Captions, respectively. The pre-trained CLIP~\cite{radford2021learning} was adopted as the video and text encoders. BertAdam was used as the optimizer, with 0.1 proportion warm-up cosine annealing, and a learning rate of 1e-4. All the models were trained for 5 epochs except on DiDeMo which was fine-tuned with 10 epochs. The temporal rank $R$ and the number of selected tokens $K$ were both set to 3. All experiments were carried out on 4 NVIDIA Tesla A100 GPUs. 

%-------------------------------------------------------------------------------------%
%-------------------------------------------------------------------------------------%
\subsection{Comparisons with State-of-the-Arts} \label{sec:4_2}
%-------------------------------------------------------------------------------------%

%------------------------------------%
\begin{table*}[t]
\renewcommand\arraystretch{1.1}
\caption{\textbf{Comparisons with state-of-the-art methods on DiDeMo, MSVD, and ActivityNet Datasets.} \faLock~ denotes using the frozen visual encoder. RAP$^{*}$ denotes the RAP model with DSL post-processing \cite{cheng2021improving}. }
\vspace{-5pt}
\resizebox{\linewidth}{!}{
%\begin{tabular}{llllllllllllllllll}
\begin{tabular}{llcccccccccccc}
\toprule%[1.5pt]
\multicolumn{2}{c}{} & \multicolumn{4}{c}{DiDeMo} & \multicolumn{4}{c}{MSVD} & \multicolumn{4}{c}{ActivityNet} \\ 
\multicolumn{1}{c}{Type} & \multicolumn{1}{l}{Methods} & \multicolumn{1}{c}{R@1↑} & \multicolumn{1}{c}{R@5↑} & \multicolumn{1}{c}{R@10↑} & \multicolumn{1}{c}{MnR↓} & \multicolumn{1}{c}{R@1↑} & \multicolumn{1}{c}{R@5↑} & \multicolumn{1}{c}{R@10↑} & \multicolumn{1}{c}{MnR↓} & \multicolumn{1}{c}{R@1↑} & \multicolumn{1}{c}{R@5↑} & \multicolumn{1}{c}{R@10↑} & \multicolumn{1}{c}{MnR↓} \\ 
\midrule
\multicolumn{1}{c}{\multirow{2}{*}{Fine-tune}} 
& \multicolumn{1}{l|} {CLIP4Clip \cite{luo2022clip4clip}} &  42.3 & 69.1 & 78.2 & \multicolumn{1}{l|} {{18.6}} &  45.5 & 75.4 & 84.1 &\multicolumn{1}{l|} {10.3}& {39.4} & {71.1} & 83.3 &{7.9} \\ 
& \multicolumn{1}{l|} 
{CLIP4Clip (\faLock~CLIP)} &  26.8 & 52.7 & 62.7 & \multicolumn{1}{l|} {{47.0}} & 36.6 &  64.5 & 73.9 &\multicolumn{1}{l|} {20.4}& {21.6} & {46.5} & 60.3 &{37.6} \\ 
\hline
\multicolumn{1}{c}{\multirow{4}{*}{Prompt}}& 
\multicolumn{1}{l|} {VPT \cite{jia2022visual}}  & 32.6 & 59.7 &  71.3 & \multicolumn{1}{l|} {30.3} & 40.8 & {69.8} & {79.8} & \multicolumn{1}{l|} {{13.7}} & 27.8 & 56.0 & 70.0 & 20.2 \\  
&\multicolumn{1}{l|} {CoOp \cite{zhou2022learning}}  & 29.7 & 56.9 &  67.9 & \multicolumn{1}{l|} {34.9} & 38.9 & 69.2 &  78.9 & \multicolumn{1}{l|} {14.0} & 29.1 & 57.3 & 72.2 & 14.2 \\

& \multicolumn{1}{l|} {{VoP$^{P}$} 
  \cite{huang2023vop}}  & 38.9 & 67.7 &  78.1 & \multicolumn{1}{l|} {{17.2}} & - & {-} & {-} & \multicolumn{1}{l|} {{-}} & 32.8 & 62.3 & 75.4 & 12.3 \\
& \multicolumn{1}{l|} {{VoP$^{C}$} 
\cite{huang2023vop}}  & 40.0 & 68.0 & 78.5 & \multicolumn{1}{l|} {18.3} & - & {-} & {-} & \multicolumn{1}{l|} {{-}} & 32.6 & 62.5 & 76.5 & 12.0 \\
\hline
\multicolumn{1}{c}{\multirow{5}{*}{Adapter}}
& \multicolumn{1}{l|}{ST-Adapter \cite{pan2022st}}  & 36.6 & 63.4 & 72.0 & \multicolumn{1}{l|} {26.7}  & 42.5 & 72.0 & 81.7 &\multicolumn{1}{l|} {12.4}& 29.8 & 59.5 & 73.7 &14.5  \\
& \multicolumn{1}{l|}{LoRA 
\cite{hu2021lora}} & 38.4 & 65.9 & 75.7 & \multicolumn{1}{l|} {22.6} & 45.1 & 75.0 & 84.0 & \multicolumn{1}{l|}{10.8}& 27.7 & 55.8 & 69.3 &18.8 \\  
& \multicolumn{1}{l|}  
{SSF \cite{lian2022scaling}} & 38.3 &  65.8 & 77.7 & \multicolumn{1}{l|} {21.8} & 43.9 & 73.3 & {82.8} & \multicolumn{1}{l|} {11.2} & 33.2 & 63.6 & 77.0 & 11.3\\
\rowcolor{aliceblue}  & \multicolumn{1}{l|} {RAP (Ours)} & {42.6} &  {70.4} &  {79.6} & \multicolumn{1}{l|} {18.0} & {44.9} & {73.7} & 83.1 &  \multicolumn{1}{l|} {{11.1}} & 40.8 & 71.0 & {82.2} & 8.3 \\
\rowcolor{aliceblue} & \multicolumn{1}{l|} {RAP$^{*}$ (Ours)} & \textbf{47.1} & \textbf{74.1} & \textbf{82.4} & \multicolumn{1}{l|}{ \textbf{13.9}} & \textbf{49.8} & \textbf{78.2} & \textbf{86.1} &\multicolumn{1}{l|}{\textbf{9.7}} & \textbf{48.4} & \textbf{76.2} & \textbf{86.4} & \textbf{7.0}
\\
\bottomrule%[1.5pt]
\end{tabular}
}
\label{tab:otherSOTA}
\end{table*}

%------------------------------------%

The comparison results are summarized in Table~\ref{tab:msrvtt} and Table~\ref{tab:otherSOTA}. Specifically, we set three sets of comparison experiments: \textbf{1)} Fine-tuning: We take the fully fine-tuned CLIP4clip \cite{luo2022clip4clip} for comparisons. Besides, we also list the zero-shot performance of CLIP4clip, \ie, \faLock~CLIP in Table~\ref{tab:msrvtt}, for comparisons; \textbf{2)} Prompt-tuning: We compare our proposed RAP to prompt-tuning methods including CoOp \cite{zhou2022learning}, VPT \cite{jia2022visual} and VoP \cite{huang2023vop}. Since VPT is tailored for purely visual tasks, we experiment by fine-tuning or freezing the textual branch of CLIP, respectively; \textbf{3)} Adapter: We conduct experiments with the state-of-the-art adapters including ST-Adapter \cite{pan2022st}, LoRA \cite{hu2021lora} and SSF \cite{lian2022scaling}. Notably, ST-Adapter is applied on the visual branch and the textual branch is either fine-tuned or freezed. For the experiments with CoOP, we insert 32 learnable prompt tokens at the input of the textual encoder.

The comparison results demonstrate the superior performance of our proposed RAP. For example, on the MSR-VTT dataset, our RAP surpasses the fully fine-tuned CLIP4clip by 2.2\% (42.6 \vs 44.8) on R@1 with only 0.7\% parameters (1.06 M \vs 151.28 M) using CLIP-ViT-B/32 backbone. Besides, we also achieve superior performance compared to current prompt-tuning and adapter-tuning methods. Although the parameters of our RAP are slightly higher than LoRA and SSF, considering the considerable performance improvement, our RAP strikes a better balance between parameters and performance. 

%------------------------------------%
\begin{table}[t]
\centering
\caption{\textbf{Comparisons of the memory footprint and GFLOPs.} The input frame number is set to 12 and the ViT-B/32 is employed as the backbone. \faUnlock~refers to the text-encoder being trainable.} 
\vspace{-5pt}
\resizebox{\linewidth}{!}{
\renewcommand\arraystretch{1.1}
\setlength{\tabcolsep}{2pt}
\begin{tabular}{lx{60}x{60}x{60}x{35}}
\toprule%[0.5pt]
Method  & \#Params (M)  & Memory (G) & GFLOPs & R@1 \\ 
\midrule 
CLIP4clip  & 151.3  &  12.9 & \textbf{54.4} & 42.6\\
ST-Adapter & 14.2  &  10.3 & 62.8 & 39.5\\
ST-Adapter \faUnlock & 77.5  & 11.2 & 62.8 & 42.5\\
LoRA  & 0.5  &  \textbf{9.5} & 67.6 & 40.5\\
SSF  & \textbf{0.3}  &  17.1 & 54.5 & 40.8\\
\rowcolor{aliceblue} RAP\_light (Ours) & 0.4 & 12.2 & 55.3 & \textbf{43.2}\\
\bottomrule
\end{tabular}
}
\label{tab:memory}
\end{table}

%------------------------------------%

Besides, to further probe the memory usage and computational complexity of the proposed model, we summarize the GPU memory usage during the training process and GFLOPs of the model in Table \ref{tab:memory}. For fair comparisons, we coequally set the number of input frames of each model to 12 frames and experiment with the ViT-B/32 backbone. We set up a lightweight RAP which only applies LoRM and ASA at the last four layers. As shown, compared to the fully fine-tuned Clip4clip, RAP\_light remarkably reduces the tunable parameters, slightly lowers the memory footprint and boosts the performance. In brief, our RAP\_light achieves the balance between computational overhead and performance, \ie, paying affordable overhead while obtaining considerable performance gains.
%-------------------------------------------------------------------------------------%
%-------------------------------------------------------------------------------------%
\subsection{Ablations Study}\label{sec:4_3}
%-------------------------------------------------------------------------------------%

We conduct all the ablation studies on the MSR-VTT dataset with the ViT-B/32 backbone. The input frame number is set to 12.

%-------------------------------------------------------------------------------------%
\noindent \textbf{Component Ablations.} We ablate the proposed low-rank modulation module and the asynchronous self-attention. The results are summarized in Table \ref{tab:componentAblate}. We can conclude that both components are crucial to superior performance at the cost of negligible parameter overhead. For example, LoRM yields a 2.3\% performance boost on R@1 with the cost of 0.42 M parameter (mode \#1 \vs mode \#3).

%-------------------------------------------------------------------------------------%
% Component Ablations   
\begin{table}[t]
\caption{\textbf{Ablations of model components of RAP.}}
\vspace{-5pt}
\label{tab:componentAblate}
\centering
\resizebox{\linewidth}{!}{
\begin{tabular}{ccccccc}
\toprule
Mode & LoRM & ASA & R@1 & R@5 & R@10 & \#Params (M) \\
\midrule
\#1  & \ding{51} & \ding{51}  & \textbf{44.8} & \textbf{71.4} & \textbf{81.5} & 1.06  \\
\#2 &  \ding{51} & \ding{55}  & 43.3 & 70.9 & 81.8 & 0.76 \\
\#3 &  \ding{55} & \ding{51}  & 42.5  & 70.1 & 80.3 & 0.64 \\
\#4 &  \ding{55} & \ding{55}  & 40.8 & 68.2 & 78.6 & \textbf{0.34}\\
\bottomrule
\end{tabular}
}
\end{table}

%-------------------------------------------------------------------------------------%
%-------------------------------------------------------------------------------------%
\begin{table}[t]
\caption{\textbf{Ablations of decomposition manners.} $\varnothing$ denotes RAP without any variants of LoRM. ``T", ``S" and ``L" represent temporal, spatial, and layer, respectively.}
\vspace{-5pt}
\label{tab:decomposeAblate}
\centering
\resizebox{0.9\linewidth}{!}{
\begin{tabular}{cccccccc}
\toprule
& Mode & R@1 & R@5 & R@10 & MdR & MnR \\
\midrule
\#1 & $\varnothing$ & 40.8 & 68.2 & 78.6 & 2.0 & 17.0\\
\#2 & T  & \textbf{43.3} & \textbf{70.9} & \textbf{81.8} & \textbf{2.0} & \textbf{14.7} \\
\#3 & S-T & 43.2 & 69.4 & 80.7 & 2.0 & 15.1 \\
\#4 & S-T-L & 42.0 & 67.8 & 80.3 & 2.0 & 14.5 \\
\bottomrule
\end{tabular}
}
\end{table}
%-------------------------------------------------------------------------------------%
%-------------------------------------------------------------------------------------%
\begin{table}[t]
\caption{\textbf{Ablations of selection manners.} $\varnothing$ indicates that none of the token selection policies is used.}
\vspace{-5pt}
\label{tab:selectAblate}
\centering
\resizebox{\linewidth}{!}{
\begin{tabular}{cccccccc}
\toprule
Mode & R@1 & R@5 & R@10 & MdR & MnR \\
\midrule
\emph{text-top-K} & \textbf{44.8} & \textbf{71.4} & \textbf{81.5} & \textbf{2.0} & \textbf{14.4} \\
\emph{text-bottom-K} & 43.0 & 70.7 & 80.3 & 2.0 & 14.8\\
\emph{vision-top-K} & 44.5 & 71.3 & 80.7 & 2.0 & 14.8 \\
\emph{vision-bottom-K} & 43.5 & 70.6 & 80.3 & 2.0 & 15.1 \\
\emph{random} & 43.2 & 70.8 & 81.2 & 2.0 & 14.9  \\
 $\varnothing$ & 41.4 & 68.9 & 79.9 & 2.0 & 15.7\\
\bottomrule
\end{tabular}
}
\end{table}
 
%-------------------------------------------------------------------------------------%
%-------------------------------------------------------------------------------------%
\begin{table}[t]
\caption{\textbf{Ablations of warping in Asynchronous attention.} T-Warp and S-Warp denote warping only on the temporal and spatial dimensions, respectively.}
\vspace{-5pt}
\label{tab:warpAblate}
\centering
\resizebox{\linewidth}{!}{
\begin{tabular}{ccccccccc}
\toprule
T-Warp  & S-Warp & R@1 & R@5 & R@10 & MdR & MnR \\
\midrule
\ding{51} & \ding{51}  & \textbf{44.8}  & \textbf{71.4}  & \textbf{81.5}  & \textbf{2.0} & \textbf{14.4} \\
\ding{55} &  \ding{51} & 44.0 & 70.4 & 81.4 &2.0 & 14.8\\
\ding{51} & \ding{55} & 44.2 & 70.9 & 81.2 & 2.0 & 14.8 \\
\bottomrule
\end{tabular}
}
\end{table}
%-------------------------------------------------------------------------------------%

\noindent \textbf{Ablations on the low-rank decompose manner of LoRM.} In Equation \eqref{eq:1}, we conduct the low-rank decomposition in the temporal dimension, and the modulation weights are with the dimension of $\mathbb{R}^{T \times D_\text{v}}$, \ie, $\mathbb{R}^{T \times D_\text{v}} \leftarrow \mathbb{R}^{T \times R} \cdot \mathbb{R}^{R \times D_\text{v}}$. Here we ablate more decomposition options: \emph{\textbf{i)}} the spatial-temporal decomposition: The modulation is applied at the spatial-temporal dimension with the weight of $\mathbb{R}^{T \times N \times D_\text{v}}$, \ie, $\mathbb{R}^{T \times N \times D_\text{v}} \leftarrow \mathbb{R}^{T \times N \times R} \cdot \mathbb{R}^{R \times D_\text{v}}$, where $T$ and $N$ denote and frame number and the patch number within each frame, respectively. \emph{\textbf{ii)}} the spatial-temporal-layer decomposition: We uniformly decompose all the modulation weights across all the layers. Specifically, the modulation weights are of the shape of $\mathbb{R}^{M \times T \times N \times D_\text{v}}$, \ie, $\mathbb{R}^{M \times T \times N \times D_\text{v}} \leftarrow \mathbb{R}^{M \times R} \cdot \mathbb{R}^{R \times T \times N \times R} \cdot \mathbb{R}^{R \times D_\text{v}}$, where $M$ denotes the inserted module number of all layers.

The comparison results are summarized in Table~\ref{tab:decomposeAblate}. From the comparison results, we observe that using temporal decomposition alone brings about the optimum performance. Additionally introducing decomposition on the spatial and layer-wise dimension leads to the performance degrade. These results manifest our motivation that video data exhibits a substantial degree of redundancy in the temporal dimension.
%-------------------------------------------------------------------------------------%

%-------------------------------------------------------------------------------------%
\noindent \textbf{Ablations on the text-conditioned selection manners.} To stabilize the training process of ASA, we propose a text-conditioned selection strategy to constrain the asynchronous attention computation within the selected top-related patch features (\cf Sec.~\ref{sec:3.3}). For clarity, we denote this filter manner as \emph{text-top-K}. Here we experiment with more visual token selection manners: \emph{\textbf{i)}} \emph{random}: randomly select $K$ patch feature within each frame; \emph{\textbf{ii)}} \emph{text-bottom-K}: For each patch token feature, we compute the sentence-patch similarity and select lowest $K$ responded patches; \emph{\textbf{iii)}} \emph{vision-top-K}: Instead of using sentence features, we compute the similarities between each patch feature and the \texttt{[CLS]} token feature of the frame. The filtered set is constituted by selecting the top K responsive patches; \emph{\textbf{iv)}} \emph{vision-bottom-K}: Similar to \emph{vision-top-K}, we compute patch-wise similarities with \texttt{[CLS]} token and select lowest $K$ responded patches; \emph{\textbf{v)}} $\varnothing$: none of the selection strategies are used and all the patch features are wrapped.
%-------------------------------------------------------------------------------------%
\begin{table}[t]
\caption{\textbf{Ablations on plug-and-play performance.} X-CLIP$^{*}$ denotes freezing the CLIP backbone of X-CLIP \cite{ma2022x}.}
\vspace{-5pt}
\label{tab:plug}
\centering
\scalebox{0.8}{
%\resizebox{\linewidth}{!}{
\begin{tabular}{ccc}
\toprule
Method & R@1↑ & \#Param (M)↓ \\
\midrule
X-CLIP \cite{ma2022x} & 46.1 & 151.3 \\
X-CLIP$^{*}$ + LoRM & 46.6 & 0.76 \\
X-CLIP$^{*}$ + LoRM + ASA & 47.9 & 1.06 \\
\bottomrule
\end{tabular}
}
\end{table}

%-------------------------------------------------------------------------------------%
%-------------------------------------------------------------------------------------%
\begin{table}[t]
	\begin{center}
		\caption{\textbf{Ablations on hyper-parameters} including the temporal rank $R$ and the number of selected token $K$.}
            \vspace{-5pt}
		\label{tab:ablaParam}
		\resizebox{0.7\linewidth}{!}{
			\begin{tabular}{cccccc}
				\toprule
				\textbf{$R$} & 1 & 3 & 5 & 7 & 9\\
				\midrule
				\textbf{R@1} & 42.6 & \textbf{44.8} & 44.0 & 43.2 & 43.0 \\
				\midrule
				\midrule
				% \bottomrule
				% \toprule
				\textbf{$K$} & 1 & 3 & 5 & 7 & 9\\
				\midrule
				\textbf{R@1} & 44.5 & \textbf{44.8} & 43.8 & 43.6 & 42.9 \\
				\bottomrule
			\end{tabular}
		}
	\end{center}
\end{table}
%-------------------------------------------------------------------------------------%
%-------------------------------------------------------------------------------------%
The comparison results of the above selection strategies are summarized in Table \ref{tab:selectAblate}. We have the following findings. Firstly, not using the token selection strategy (\ie, $\varnothing$ in Table \ref{tab:selectAblate}) causes substantial performance degradation, \eg, reaching only 41.4\% on R@1. This is probably because warping each patch tokens wreaks havoc on the well-trained CLIP weights. Secondly, our proposed \emph{text-top-K} policy outperforms the other ones on all five metrics. This demonstrates that selectively warping partial patch tokens in a parameterized way can better adapt the vanilla CLIP to the video scenario.
%-------------------------------------------------------------------------------------%

%-------------------------------------------------------------------------------------%
\noindent \textbf{Ablations on the warping manner of ASA.} In Sec.~\ref{sec:3.3}, we predict the patch-wise warping distance in both the temporal and spatial dimensions. Here we ablate either of the two dimensions to see the difference. As shown in Table \ref{tab:warpAblate}, restricting warping in either temporal or spatial dimension will lead to performance degradation, which demonstrates that free-form patch-wise warping is crucial to the final performance.
%-------------------------------------------------------------------------------------%

%-------------------------------------------------------------------------------------%
\noindent \textbf{Ablations on plug-and-play performance.} Both the proposed LoRM and ASA modules serve as plug-and-play modules and can be compatible with versatile CLIP-based methods. To demonstrate this, we conduct experiments based on the more advanced CLIP-based method X-CLIP \cite{ma2022x}. Specifically, we freeze the CLIP backbone and then insert LoRM within each Transformer layer and replace the vanilla self-attention with our proposed ASA. The comparison results in Table \ref{tab:plug} show that our LoRM and ASA can consistently benefit the retrieval performance even with the more advanced X-CLIP baseline. Besides, compared to the fully finetuned counterpart, our proposed LoRM and ASA show great advantages in terms of tunable parameters.
%-------------------------------------------------------------------------------------%

%-------------------------------------------------------------------------------------%
\noindent \textbf{Ablations on hyper-parameters.} We conduct ablation studies on the temporal rank $R$ and selected token number $K$ in Table \ref{tab:ablaParam}. We set $R=3$ and $K=3$ to achieve the best retrieval performance.
%-------------------------------------------------------------------------------------%

%-------------------------------------------------------------------------------------%
%\input{exps/4_4_vis}
%-------------------------------------------------------------------------------------%
\section{Conclusions}

In this work, we present RAP to efficiently transfer the pre-trained CLIP model to TVR. To accommodate the inherent video structure and the cross-modality setting, we introduce a low-rank modulation module to achieve the frame-wise sparse representation and an asynchronous self-attention module to enhance the cross-frame correlations. Extensive experiments illustrate that RAP achieves comparable or even better performance than previous arts and the fully fine-tuned counterpart. 

\newcommand{\myparagraph}[1]{\textbf{#1}\hspace{1.8ex}}
\section*{Impact Statements}
\noindent \myparagraph{Ethics Statement.} Our RAP aims to conduct parameter-efficient text-video retrieval through a temporally sparse and correlated adapter. The ethical issues may exist in the following two perspectives. Firstly, similar to many data-driven methods, there are concerns about the issue of data privacy, anonymization, and compliance with relevant data protection regulations. Secondly, the considerations related to potential bias in the dataset and the retrieval model, especially concerning sensitive topics, should be acknowledged. We are transparent about the ethical considerations in our research to uphold the integrity of the academic process and to ensure that this work aligns with ethical standards and norms in the field.

\noindent \myparagraph{Limitation.} Despite the remarkable progress, our RAP still faces several limitations. Firstly, we use the text-conditioned selection to filter the most representative visual patches. Due to the semantic gap conveyed by the textual and visual signals, the alignment of complex concepts and contexts across different modalities should be conducted in a more fine-grained manner. Secondly, due to the limitations of computing resources, we experiment with the backbone of ViT-B/32 and ViT-B/16. The salable experiments on ViT-L/14 and ViT-E/14 backbones are left for future work. 
\section*{Acknowledgements}

This work was supported by the Natural Science Foundation of China (No.62202014, 62332002), Shenzhen Basic Research Program under Grant JCYJ20220813151736001

\bibliography{main}

\clearpage
\appendix
%-----------------------------------------------------------------------
\section{Appendix} \label{sec:appendix}
%-----------------------------------------------------------------------

%-----------------------------------------------------------------------
\section*{A.  More Experiments}
%-----------------------------------------------------------------------

%-----------------------------------------------------------------------
\newcommand{\tabfootnotesize}{\fontsize{8}{9}\selectfont}

\makeatletter
  \newcommand\figcaption{\def\@captype{figure}\caption}
  \newcommand\tabcaption{\def\@captype{table}\caption}
\makeatother

%\vspace{8pt}
\begin{minipage}[b]{\textwidth}
\tabcaption{\textbf{Video-to-text retrieval results on DiDeMo, MSVD, and ActivityNet Datasets.} \faLock~ denotes using the frozen visual encoder.}
\vspace{-5pt}
\centering
\resizebox{1.0\linewidth}{!}{
\begin{tabular}{llcccccccccccc}
\toprule%[1.5pt]
\multicolumn{2}{c}{} & \multicolumn{4}{c}{DiDeMo} & \multicolumn{4}{c}{MSVD} & \multicolumn{4}{c}{ActivityNet} \\ 
\multicolumn{1}{c}{Type} & \multicolumn{1}{l}{Methods} & \multicolumn{1}{c}{R@1↑} & \multicolumn{1}{c}{R@5↑} & \multicolumn{1}{c}{R@10↑} & \multicolumn{1}{c}{MnR↓} & \multicolumn{1}{c}{R@1↑} & \multicolumn{1}{c}{R@5↑} & \multicolumn{1}{c}{R@10↑} & \multicolumn{1}{c}{MnR↓} & \multicolumn{1}{c}{R@1↑} & \multicolumn{1}{c}{R@5↑} & \multicolumn{1}{c}{R@10↑} & \multicolumn{1}{c}{MnR↓} \\ 
\midrule
\multicolumn{1}{c}{\multirow{2}{*}{Fine-tune}} 
& \multicolumn{1}{l|} {CLIP4Clip \cite{luo2022clip4clip}} &  42.2 & 70.3 & 79.3 & \multicolumn{1}{l|} {{11.8}} &  56.6 & 79.7 & 84.3 &\multicolumn{1}{l|} {7.6}& {41.9} & {72.2} & 84.6 &{7.3} \\ 
& \multicolumn{1}{l|} 
{CLIP4Clip (\faLock~CLIP)} &  20.2 & 44.2 & 55.0 & \multicolumn{1}{l|} {{43.1}} & 56.3 &  82.6 & 89.8 &\multicolumn{1}{l|} {4.8}& {17.7} & {40.7} & 54.1 &{42.5} \\ 
\hline
\multicolumn{1}{c}{\multirow{4}{*}{Prompt}}& 
\multicolumn{1}{l|} {VPT \cite{jia2022visual}}  & 33.1 & 59.8 &  69.9 & \multicolumn{1}{l|} {22.7} & 59.5 & {82.9} & {88.8} & \multicolumn{1}{l|} {{5.9}} & 28.4 & 56.7 & 69.4 & 19.7 \\  
&\multicolumn{1}{l|} {CoOp \cite{zhou2022learning}}  & 32.3 & 57.0 &  68.2 & \multicolumn{1}{l|} {23.4} & 53.8 & 78.3 &  82.9 & \multicolumn{1}{l|} {12.4} & 29.0 & 57.6 & 72.4 & 14.0 \\
& \multicolumn{1}{l|} {{VoP$^{P}$} 
  \cite{huang2023vop}}  & 40.6 & 68.3 & 78.6 & \multicolumn{1}{l|} {11.6} & - & {-} & {-} & \multicolumn{1}{l|} {{-}} & 34.8 & 65.0 & 78.2 & 10.7 \\
& \multicolumn{1}{l|} {{VoP$^{C}$} 
\cite{huang2023vop}}  & 39.1 & 65.3 & 76.7 & \multicolumn{1}{l|} {13.8} & - & {-} & {-} & \multicolumn{1}{l|} {{-}} & 34.2 & 64.8 & 78.4 & 10.7 \\
\hline
\multicolumn{1}{c}{\multirow{5}{*}{Adapter}}
& \multicolumn{1}{l|}{ST-Adapter \cite{pan2022st}}  & 35.9 & 61.0 & 72.0 & \multicolumn{1}{l|} {20.1}  & 53.6 & 80.5 & 87.3 &\multicolumn{1}{l|} {5.8}& 29.7 & 58.8 & 73.1 & 15.5  \\ 
& \multicolumn{1}{l|}{LoRA 
\cite{hu2021lora}} & 39.7 & 66.8 & 77.3 & \multicolumn{1}{l|} {13.9} & 64.3 & 87.3 & 92.5 & \multicolumn{1}{l|}{4.1}& 30.8 & 60.0 & 73.2 & 15.2 \\  
& \multicolumn{1}{l|}
{SSF \cite{lian2022scaling}} & 40.0 &  67.1 & 77.4 & \multicolumn{1}{l|} {13.2} & 61.9 & 87.0 & {90.7} & \multicolumn{1}{l|} {4.5} & 36.2 & 66.9 & 79.0 & 10.4\\
\rowcolor{aliceblue}  & \multicolumn{1}{l|} {RAP (Ours)} & {44.0} &  69.2 &  {80.1} & \multicolumn{1}{l|} {10.5} & 65.2 & 88.7 & 93.1 &  \multicolumn{1}{l|} {4.2} & 41.9 & 73.0 & {84.0} & 7.5 \\
\rowcolor{aliceblue} & \multicolumn{1}{l|} {RAP$^{*}$ (Ours)} & \textbf{47.7} & \textbf{74.4} & \textbf{83.2} & \multicolumn{1}{l|}{ \textbf{9.5}} & \textbf{69.6} & \textbf{91.9} & \textbf{95.7} &\multicolumn{1}{l|}{\textbf{2.9}} & \textbf{48.2} & \textbf{76.5} & \textbf{86.2} & \textbf{6.8}
\\
\bottomrule
\end{tabular}
}
\label{tab:v2totherSOTA}
% \end{table}
\end{minipage}
%-----------------------------------------------------------------------

%-----------------------------------------------------------------------
\noindent \textbf{Video-to-text Performance:} We supplement the video-to-text performance on the DiDeMo, MSVD and ActivityNet Captions datasets in Table \ref{tab:v2totherSOTA}. The experimental results consistently demonstrate the superiority of our RAP. For example on the MSVD dataset, RAP surpasses the fully fine-tuned method by 8.6\% on R@1.
%-----------------------------------------------------------------------

%-----------------------------------------------------------------------
\noindent \textbf{Low-rank modulation on textual features:} In Sec.~\ref{sec:3.2}, we apply the low-rank decomposition modulation in the visual branch, specifically in the temporal dimension. Here we apply the low-rank modulation on the textual branch to see the differences. Concretely, the modulation weights are with the shape of $\mathbb{R}^{W \times D_\text{t}} \leftarrow \mathbb{R}^{W \times R_\text{t}} \cdot \mathbb{R}^{R_\text{t} \times D_\text{t}}$, where $W$ denotes the total word length, $R_\text{t}$ is the low-rank hyper-parameter and $D_\text{t}$ is the textual feature dimension. We set $R_\text{t} = 3$. 

The ablation results are shown in Table \ref{tab:text_low_rank}. As shown, applying the low-rank modulation on textual features causes performance degradation, which may be because word-level representations do not exhibit the same redundancy as frame-level visual features.
%-----------------------------------------------------------------------

%-----------------------------------------------------------------------
\vspace{8pt}
%\vspace{20cm}
\begin{minipage}[b]{0.45\textwidth}
\tabcaption{\textbf{Ablations of the low-rank modulation on the textual branch.}}
\vspace{-5pt}
\centering
\resizebox{1.0\linewidth}{!}{
\begin{tabular}{ccccccccc}
\toprule
LoRM on Text & R@1 & R@5 & R@10 & MnR & \#Param\\
\midrule
\ding{55}  & \textbf{44.8}  & \textbf{71.4}  & \textbf{81.5}  & \textbf{14.4} & \textbf{1.06M} \\
\ding{51} & 44.3 & 72.1 & 81.0 & 14.4 & 1.48M\\
%\emph{frame-wise} & \ding{51} & ---  & 44.4 & 70.8  & 81.1 & 2.0& 14.9 \\
\bottomrule
\end{tabular}
}
\label{tab:text_low_rank}
% \end{table}
\end{minipage}

%-----------------------------------------------------------------------

~\\ 
~\\ 
~\\ 
~\\ 
~\\ 
~\\ 
~\\ 
~\\ 
~\\ 
~\\ 
~\\ 
~\\ 
~\\ 
~\\ 
~\\ 
~\\ 
~\\ 
~\\ 
~\\ 
~\\ 
~\\ 
~\\ 

%-----------------------------------------------------------------------
\section*{B. Visualization Results}
%-----------------------------------------------------------------------

\begin{figure*}[t]
	\centering
	\includegraphics[width=0.85\textwidth]{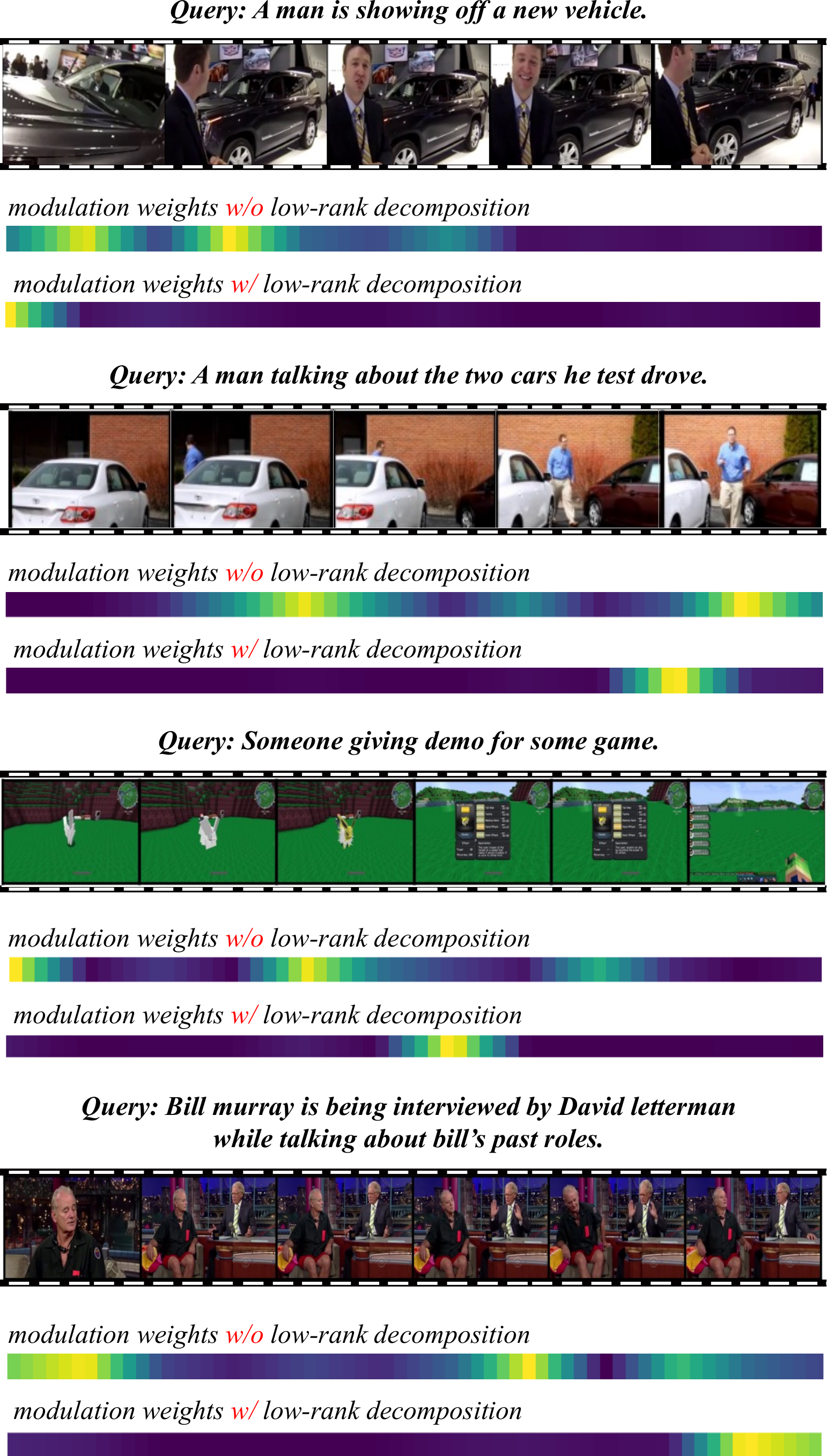}
	\caption{\textbf{Illustrations of temporal sparsity.} We visualize the modulation weight \emph{w/} or \emph{w/o} low-rank decomposition.}
	\label{fig:vis_sparse}
	%\vspace{-2mm}
\end{figure*}

%-----------------------------------------------------------------------
We visualize the per-frame modulation weights with or without the low-rank decomposition. As shown in Figure \ref{fig:vis_sparse}, the modulation weights with decomposition demonstrate more salient distributions, which manifests the temporal sparsity characteristic of video data.

\begin{figure*}[t]
	\centering
	\includegraphics[width=0.85\textwidth]{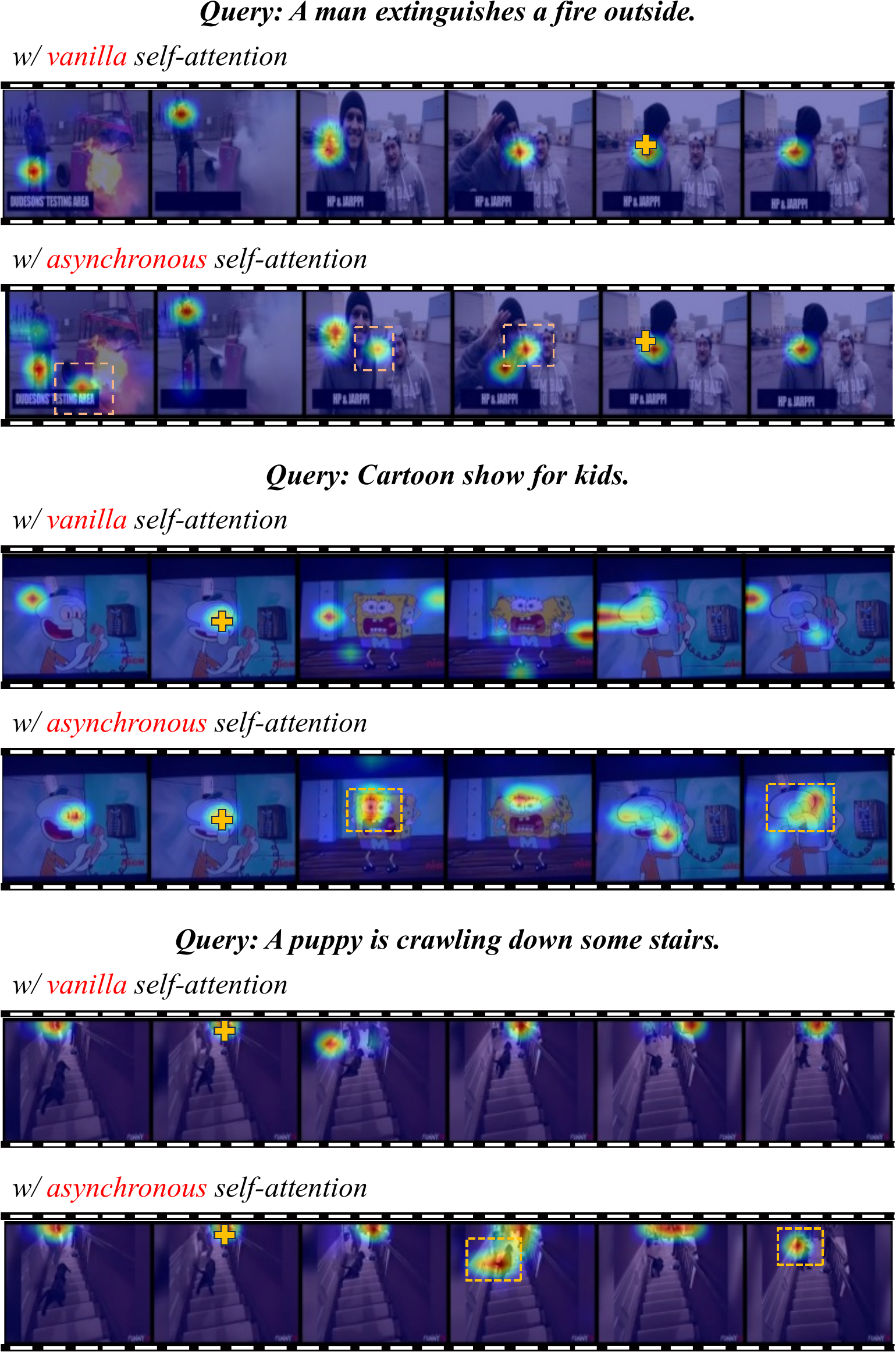}
	\caption{\textbf{Illustrations of temporal correlation.} The query patch is marked by the yellow cross and the similarity map within other frames are plotted.}
	\label{fig:vis_correlate}
	%\vspace{-2mm}
\end{figure*}

Besides, we visualize the effect of the asynchronous self-attention. We randomly select one patch feature (\inlineMarker) in the frame and compute its similarity distribution with the patches in other frames. The results in Figure \ref{fig:vis_correlate} show that the proposed asynchronous self-attention can adaptively attend to patch-of-interest, which leads to effective temporal correlation modeling.

%-----------------------------------------------------------------------
%-----------------------------------------------------------------------
%-----------------------------------------------------------------------

%-----------------------------------------------------------------------

\end{document}